\documentclass[runningheads,a4paper]{llncs}

\usepackage{amssymb}
\usepackage{graphicx}
\usepackage{marvosym}
\usepackage{times}
\usepackage{helvet}
\usepackage{courier}
\usepackage{amsmath}
\usepackage{multirow}
\usepackage{amsopn}
\usepackage{bm}
\usepackage{epsfig}
\usepackage{subfigure}
\usepackage{dsfont}
\usepackage{epstopdf}
\usepackage{amssymb}
\usepackage{epsfig}
\usepackage{graphicx,subfigure}
\usepackage{enumerate}
\usepackage{amscd}
\usepackage{rotating}
\usepackage{algorithm,algorithmic}
\usepackage[usenames,dvipsnames]{color}
\usepackage{color}
\usepackage{colortbl}
\definecolor{mygray}{gray}{.9}
\definecolor{mypink}{rgb}{.99,.91,.95}
\definecolor{mycyan}{cmyk}{.3,0,0,0}

\usepackage{url}

\urldef{\mailsa}\path|ducyatict@gmail.com  |
\urldef{\mailsb}\path|{changying, guoping}@iscas.ac.cn |
\begin{document}

\mainmatter  

\title{Efficient and Adaptive Kernelization for Nonlinear Max-margin Multi-view Learning}

\titlerunning{Efficient and Adaptive Kernelization for Nonlinear Max-margin Multi-view Learning}

%
%
\author{Changying Du \inst{1,3} \Letter 
\and Jia He \inst{2,5}
\and Changde Du \inst{4,5}
\and Fuzhen Zhuang \inst{2}
\and Qing He \inst{2,5}
\and  Guoping Long \inst{3} }
\authorrunning{C. Du et al.}

\institute{Huawei Noah's Ark Lab, Beijing 100085, China
\and Key Laboratory of Intelligent Information Processing, Institute of Computing Technology, Chinese Academy of Sciences, Beijing 100190, China
\and
Institute of Software, Chinese Academy of Sciences, Beijing 100190, China
\and
Research Center for Brain-Inspired Intelligence, Institute of Automation, CAS, Beijing, China
\and
University of Chinese Academy of Sciences, Beijing 100049, China\\
\mailsa}

%
%

\maketitle

\begin{abstract}
Existing multi-view learning methods based on kernel function either require the user to select and tune a single predefined kernel or have to compute and store many Gram matrices to perform multiple kernel learning. Apart from the huge consumption of manpower, computation and memory resources, most of these models seek point estimation of their parameters, and are prone to overfitting to small training data.  This paper presents an adaptive kernel nonlinear max-margin multi-view learning model under the Bayesian framework. Specifically, we regularize the posterior of an efficient multi-view latent variable model by explicitly mapping the latent representations extracted from multiple data views to a random Fourier feature space where max-margin classification constraints are imposed. Assuming these random features are drawn from Dirichlet process Gaussian mixtures, we can adaptively learn shift-invariant kernels from data according to Bochners theorem. For inference, we employ the data augmentation idea for hinge loss, and design an efficient gradient-based MCMC sampler in the augmented space. Having no need to compute the Gram matrix, our algorithm scales linearly with the size of training set. Extensive experiments on real-world datasets demonstrate that our method has superior performance.
\end{abstract}

\section{Introduction}
Nowadays data typically can be collected from various information channels, e.g., a piece of news is consisted of text, audio, video clip and hyperlink. How to effectively and efficiently fuse the information from different channels for specific learning tasks is a problem (known as multi-view learning) attracting more and more attention. For supervised learning, e.g., news classification, we have not only these multi-channel features but also the corresponding category labels. Making full use of the label information usually is critical for the construction of predictive multi-view models since it can help to learn more discriminative features for classification. A popular choice is to exploit the maximum margin principle to guarantee the learned model to have a good generalization ability~\cite{Xu2014Large,Du2015Bayesian,heonline}. Though improved performance is reported on many problems, these methods have made linearity assumption on the data, which may be inappropriate for multi-view data revealing nonlinearities.

Kernel method~\cite{Hofmann2007Kernel} is a principled way for introducing nonlinearity into linear models, and a kernel machine can approximate any function or decision boundary arbitrarily well by tuning its kernel parameter. Along this line, many single kernel multi-view learning methods have been proposed~\cite{farquhar2005two,Szedmak2007Synthesis,Fang2012Simultaneously,sun2013multi,quang2013unifying,Xu2014Large},
which typically require the user to select and tune a predefined kernel for every view. Choosing an appropriate kernel for real-world situations is usually not easy for users without enough domain knowledge. The performance of these models may be greatly affected by the choice of kernel. An alternative solution to resolve this problem is provided by multiple kernel learning (MKL), which can predefine different kernels for each data view and then integrate the kernels by algorithms such as semi-definite programming (SDP)~\cite{lanckriet2004learning}, semi-infinite linear programming (SILP)~\cite{Sonnenburg2006Large}, and simple MKL~\cite{rakotomamonjy2008simplemkl}.
However, MKL models inherently have to compute and store many Gram matrices to get good performance while computing a Gram matrix for a data set with $N$ instances and $D$ features needs $O(N^2D)$ operations and storing too many Gram matrices often leads to out of memory on commonly used computers.
Besides, all the aforementioned kernelized models seek single point estimation of their parameters, thus are prone to overfitting to small training data.
Under the Bayesian framework, BEMKL~\cite{Gonen2012Bayesian} is a state-of-the-art variational method that estimates the entire posterior distribution of model weights. Unfortunately, BEMKL has to perform time-consuming matrix inversions to compute the posterior covariances of sample and kernel weights.
Moreover, BEMKL's performance may be limited by its mean-field assumption on the approximate posterior and the absence of max-margin principle. To improve BEMKL, \cite{Du2016Efficient} proposed an efficient MCMC algorithm in the augmented variable space, however, it was still limited by the memory efficiency issue. To improve both time and memory efficiencies, \cite{Du2016Online} proposed an online Bayesian MKL model, which still relied on predefined kernels.

To address the aforementioned problems, we propose an adaptive kernel maximum margin multi-view learning (M$^3$L) model with low computational complexity. Specifically, we firstly propose an efficient multi-view latent variable model (LVM) based on the traditional Bayesian Canonical Correlation Analysis (BCCA)~\cite{wang2007variational}, which learns the shared latent representations for all views. To adaptively learn the kernel, we introduce the random Fourier features which constructs an approximate primal space to estimate kernel evaluations $K(x,x')$ as the dot product of finite vectors $\varphi(x)^{T}\varphi(x')$~\cite{rahimi2007random}. Assuming these random features are drawn from Dirichlet process (DP) Gaussian mixtures, we can adaptively learn shift-invariant kernels from data according to Bochners theorem. With such a stochastic random frequency distribution, it is more general than the traditional kernel method approach with a fixed kernel. Secondly, to make full use of the label information, we impose max-margin classification constraints on the explicit expressions $\varphi(x)$ in random Fourier feature space. Next, we regularize the posterior of the efficient multi-view LVM by explicitly mapping the latent representations to a random Fourier feature space where max-margin classification constraints are imposed. Our model is based on the data augmentation idea for max-margin learning in the Bayesian framework, which allows us automatically infer parameters of adaptive kernel and the penalty parameter of max-margin learning model. For inference, we devise a hybrid Markov chain Monte Carlo (MCMC) sampler. To infer random frequencies, we use an effective distributed DP mixture models (DPMM)~\cite{ge2015distributed}. Moreover, some key parameters don't have corresponding conjugate priors. So we adopt the gradient-based MCMC sampler Hamiltonian Monte Carlo (HMC)~\cite{neal2011mcmc} for faster convergence. The computational complexity of our algorithm is linear w.r.t. the number of instances $N$. Extensive experiments on real-world datasets demonstrate our method has a superior performance, compared with a number of competitors.

\vspace{-0.2cm}
\section{Model}
\vspace{-0.2cm}
BCCA~\cite{wang2007variational} assumes the following generative process to learn the shared latent representations from multiple views $\{\mathrm{\mathbf{x}}_i\}_{i=1}^{N_v}$, where $N_v$ is the number of views:
\begin{displaymath}
\begin{array}{rcl}
\vspace{0.1cm}
\mathrm{\mathbf{h}} & \sim & \mathcal{N}(\mathrm{\mathbf{0}}, \ \mathrm{\mathbf{I}})\\
\mathrm{\mathbf{x}}_i & \sim &
\mathcal{N}(\mathrm{\mathbf{W}}_i{\mathrm{\mathbf{h}}},\mathrm{\mathbf{\Psi}}_i),
\end{array}
\end{displaymath}where $\mathcal{N}(\cdot)$ denotes the normal distribution, ${\mathrm{\mathbf{h}}} \in \mathbb{R}^{m}$ is the shared latent variable, $\mathrm{\mathbf{W}}_i \in \mathbb{R}^{D_i \times m}$ is the linear transformation and $\mathrm{\mathbf{\Psi}}_i \in \mathbb{R}^{D_i \times D_i}$ denotes covariance matrix. A commonly used prior is the Wishart distribution for $\mathrm{\mathbf{\Psi}}_i^{-1}$.

\vspace{-0.2cm}
\subsection{An Efficient Multi-view LVM}
However, BCCA has to perform time-consuming inversions of the high-dimension covariance matrix $\mathrm{\mathbf{\Psi}}_i$ which could result in a severe computational problem. To solve the problem, we propose an efficient model by introducing additional latent variables. We assume the following generative process:
\begin{displaymath}
\begin{array}{rcl}
\vspace{0.1cm}
\mathrm{\mathbf{h}}, \ \mathrm{\mathbf{u}}_i & \sim & \mathcal{N}(\mathrm{\mathbf{0}}, \ \mathrm{\mathbf{I}})\\
\mathrm{\mathbf{x}}_i & \sim &
\mathcal{N}(\mathrm{\mathbf{W}}_{i}{\mathrm{\mathbf{h}}}+\mathrm{\mathbf{V}}_{i}{\mathrm{\mathbf{u}}}_i,{\tau}_i^{-1}\mathrm{\mathbf{I}})
,\end{array}
\end{displaymath}where ${\mathrm{\mathbf{u}}}_i \in \mathbb{R}^{K_i}$ is the additional latent variable. Gamma prior can be used for $\tau_i$ and popular automatic relevance determination (ARD) prior can be imposed on projection matrices $\mathrm{\mathbf{W}}_i$,\ $\mathrm{\mathbf{V}}_{i} \in \mathbb{R}^{D_i \times K_i}$, i.e.,
\begin{displaymath}
\begin{array}{rcl}
\vspace{0.2cm}
{\mathrm{\mathbf{r}}}_i & \sim  & \prod_{j=1}^{m}\Gamma(r_{ij}|a_{r},b_{r})\\ \vspace{0.15cm}
\mathrm{\mathbf{W}}_{i} & \sim  &
\prod_{j=1}^{m}\mathcal{N}(\mathrm{\mathbf{w}}_{i,\cdot j}|\mathrm{\mathbf{0}}, r^{-1}_{ij}\mathrm{\mathbf{I}})\\ \vspace{0.15cm}
\mathrm{\mathbf{V}}_{i} & \sim  &
\prod_{j=1}^{K_i}\mathcal{N}(\mathrm{\mathbf{v}}_{i,\cdot j}|\mathrm{\mathbf{0}}, \ \rm {\eta}^{-1}\mathrm{\mathbf{I}})\\ \vspace{0.15cm}
\tau_i & \sim  &
\Gamma(\tau_i|a_{\tau},b_{\tau})
,\vspace{-0.15cm}\end{array}
\end{displaymath}where $i=1,\cdots,N_v$, $\Gamma(\cdot)$ denotes Gamma distribution, $\mathrm{\mathbf{w}}_{i,\cdot j}$ represents the $j$-th column of the transformation matrix $\mathrm{\mathbf{W}}_{i}$ and $\mathrm{\mathbf{v}}_{i,\cdot j}$ represents the $j$-th column of $\mathrm{\mathbf{V}}_{i}$. This model can be shown to be equivalent to imposing a low-rank assumption $\mathrm{\mathbf{\Psi}}_i=\mathrm{\mathbf{V}}_{i}\mathrm{\mathbf{V}}_{i}^{\mathrm{ T }}+{\tau}_i^{-1}\mathrm{\mathbf{I}}$ for the covariances, which allows decreasing the computational complexity.

We define that the data matrix of the $i$-th view is ${\mathrm{\mathbf{X}}}_i \in \mathbb{R}^{D_i \times N}$ consisting of $N$ observations $\{{\mathrm{\mathbf{x}}_i^n}\}_{n=1}^{N}$, $\widetilde{{\mathrm{\mathbf{X}}}}=\{{\mathrm{\mathbf{X}}}_i\}_{i=1}^{N_v}$ , ${\mathrm{\mathbf{H}}}=\{ {\mathrm{\mathbf{h}}}^n \}_{n=1}^N$ and ${\mathrm{\mathbf{U}}}_i=\{ {\mathrm{\mathbf{u}}}_i^n \}_{n=1}^N$. For simplicity, let $\Omega=({\mathrm{\mathbf{r}}}_i, {\mathrm{\mathbf{V}}}_i, \mathrm{\mathbf{W}}_i, \eta, \tau_i, {\mathrm{\mathbf{H}}},{\mathrm{\mathbf{U}}}_i)$ be the parameters of the multi-view LVM and $p_0(\Omega)$ be the prior of $\Omega$. We can verify that the Bayesian posterior distribution $p(\Omega|\widetilde{{\mathrm{\mathbf{X}}}})=p_0(\Omega)p(\widetilde{{\mathrm{\mathbf{X}}}}|\Omega)/p(\widetilde{{\mathrm{\mathbf{X}}}})$ can be equivalently obtained by solving the following optimization problem:
\begin{equation*}
\min_{q(\Omega)\in \mathcal{P}} \text{KL}(q(\Omega)\|p_0(\Omega))-\mathbb{E}_{q(\Omega)}[\text{log}p(\widetilde{{\mathrm{\mathbf{X}}}}|\Omega)]
,\end{equation*}where $\text{KL}(q\|p)$ is the Kullback-Leibler divergence, and $\mathcal{P}$ is the space of probability distributions. When the observations are given, $p(\widetilde{{\mathrm{\mathbf{X}}}})$ is a constant.

\vspace{-0.1cm}
\subsection{Adaptive Kernel M$^3$L}
From the description above, we can see that the multi-view LVM is an unsupervised model which learns the shared latent variables from the observations without using any label information. In general, we prefer that the shared latent representation can not only explain the observed data well but also help to learn a predictive model, which predicts the responses of new observations as accurate as possible.

Moreover, this multi-view LVM is a linear multi-view representation learning algorithm, but multi-view data usually reveal nonlinearities in many scenarios of real-world. As is well known, kernel methods are attractive because they can approximate any function or decision boundary arbitrarily well. So in this section we propose an adaptive kernel max-margin multi-view learning (M$^3$L) model. With the method named random features for the approximation of kernels~\cite{rahimi2007random}, we can get explicit expression of the latent variable ${\mathrm{\mathbf{h}}}$ in random feature space. Then we can classify these explicit expression linearly by introducing the max-margin principle which has good generalization performance. To incorporate the nonlinear max-margin method to the unsupervised multi-view LVM, we adopt the posterior regularization strategy~\cite{jaakkola1999maximum,zhu2012medlda}. Suppose we have a $1\times N$ label vector $\mathrm{\mathbf{y}}$ with its element $y^n\in \{+1,-1\},\ n=1,\cdots,N$. Then we define the following pseudo-likelihood function of latent representation ${\mathrm{\mathbf{h}}}^n$ for the $n$-th observation $\{ \mathrm{\mathbf{x}}_i^n\}_{i=1}^{N_v}$:
\begin{gather*}
\ell(y^n|\tilde{\varphi}(\mathrm{\mathbf{h}}^n),\boldsymbol{\beta})  =\exp\{-2C\cdot \max(0,1 - y^n \boldsymbol{\beta}^{\mathrm{ T }}\tilde{\varphi}(\mathrm{\mathbf{h}}^n))\},\\
{\varphi}(\mathrm{\mathbf{h}}^n) = \frac{1}{\sqrt{M}}[\cos(\bm{\omega}_1^{\mathrm{ T }}\mathrm{\mathbf{h}}^n),\cdots,\cos(\bm{\omega}_M^{\mathrm{ T }}\mathrm{\mathbf{h}}^n), \sin(\bm{\omega}_1^{\mathrm{ T }}\mathrm{\mathbf{h}}^n),\cdots,\sin(\bm{\omega}_M^{\mathrm{ T }}\mathrm{\mathbf{h}}^n)]^{\mathrm{ T }}
,\end{gather*}where $C$ is the regularization parameter, $\tilde{\varphi}(\mathrm{\mathbf{h}}^n)$ is the explicit expression of the latent variable $\mathrm{\mathbf{h}}^n$ in random Fourier feature space and $\tilde{\varphi}(\mathrm{\mathbf{h}}^n)=({{\varphi}(\mathrm{\mathbf{h}}^n)}^{\mathrm{ T }},1)^{\mathrm{ T }}$. ${\bm{\omega}}_i \in \mathbb{R}^{m}$ denotes random frequency vector and $\boldsymbol{\beta}^{\mathrm{ T }}\tilde{\varphi}(\mathrm{\mathbf{h}}^n)$ is a discrimination function parameterized by $\boldsymbol{\beta} \in \mathbb{R}^{2M+1}$.

Bochners theorem states that a continuous shift-invariant kernel  $K(\mathrm{\mathbf{h}},\bar{\mathrm{\mathbf{h}}})=k(\mathrm{\mathbf{h}}-\bar{\mathrm{\mathbf{h}}})$ is a positive definite function if and only if $k(t)$ is the Fourier transform of a non-negative measure $\rho(\bm{\omega})$ ~\cite{rudin2011fourier}. Further, we note that if $k(0)=1$, $\rho(\bm{\omega})$ will be a normalized density. So we can get
\begin{align}
\nonumber  k(\mathrm{\mathbf{h}}-\bar{\mathrm{\mathbf{h}}}) & =  \int_{\mathbb{R}^{m}} \rho({\bm{\omega}})\exp(i{\bm{\omega}}^{\mathrm{ T }}(\mathrm{\mathbf{h}}-\bar{\mathrm{\mathbf{h}}}))d{\bm{\omega}}
 =  \mathbb{E}_{{\bm{\omega}}\sim\rho}[\exp(i{\bm{\omega}}^{\mathrm{ T }}\mathrm{\mathbf{h}})\exp(i{\bm{\omega}}^{\mathrm{ T }}\bar{\mathrm{\mathbf{h}}})^{*}]\\
\nonumber  & \approx  \frac{1}{M}\sum_{j=1}^{M} \exp(i{\bm{\omega}}_j^{\mathrm{ T }}\mathrm{\mathbf{h}})\exp(i{\bm{\omega}}_j^{\mathrm{ T }}\bar{\mathrm{\mathbf{h}}})^{*}.
\end{align}If the kernel $k$ is real-valued, we can discard the imaginary part:
\begin{gather*}
k(\mathrm{\mathbf{h}}-\bar{\mathrm{\mathbf{h}}})  \approx  \varphi(\mathrm{\mathbf{h}})^{\mathrm{ T }}\varphi(\bar{\mathrm{\mathbf{h}}})\\
\varphi(\mathrm{\mathbf{h}})  \equiv  \frac{1}{\sqrt{M}}[\cos({\bm{\omega}}_1^{\mathrm{ T }}\mathrm{\mathbf{h}}),\cdots,\cos({\bm{\omega}}_M^{\mathrm{ T }}\mathrm{\mathbf{h}}), \sin({\bm{\omega}}_1^{\mathrm{ T }}\mathrm{\mathbf{h}}),\cdots,\sin({\bm{\omega}}_M^{\mathrm{ T }}\mathrm{\mathbf{h}})]^{\mathrm{ T }}.
\end{gather*}
A robust and flexible choice of  $\rho(\bm{\omega})$ is a Gaussian mixture model. Mixture models based on DPs treat the number of represented mixture components as a latent variable, and infer it automatically from observed data. DP Gaussian mixture prior is widely used for density estimation~\cite{Oliva2016Bayesian}. Assuming these random features are drawn from DP Gaussian mixtures, we can adaptively learn shift-invariant kernels from data according to Bochners theorem. We impose DP Gaussian mixture prior for the variables ${\bm{\omega}}_j,j=1,\cdots,M$. Suppose the DP has base distribution $G_0$ and concentration parameter ${\alpha}$, then we have
\begin{align}
\nonumber {\zeta}_k \sim G_0, \ \   {\nu}_k\sim \text{Beta}(1,{\alpha}), \ \  {\varpi}_k={\nu}_k\prod_{i=1}^{k-1}(1-{\nu}_i)\\
\nonumber z_j \sim \text{Cat}({\bm{\varpi}}), \quad  {\bm{\omega}}_j \sim \mathcal{N}({\zeta}_{z_j}), \quad \quad \quad \quad
\end{align}
where $k=1,\cdots,\infty$, and ${\zeta}_{k}=({\bm{\mu}}_k, {\bm{\Sigma}}_k)$ contains the mean and covariance parameters of the $k$-th Gaussian component. Popular choice would be Normal-Inverse-Wishart prior $G_0$ for the mixture components:
\begin{align}
\nonumber \bm{\Sigma}_k \sim \mathcal{W}^{-1}(\bm{\Psi}_0,\nu_0), \ \ \ \  \bm{\mu}_k \sim \mathcal{N}(\bm{\mu}_0,\frac{1}{\kappa_0}{\bm{\Sigma}}_k),
\end{align}where $\mathcal{W}^{-1}(\cdot)$ denotes Inverse-Wishart distribution. Next, we impose prior on $\bm{\beta}$ as the following form
\begin{displaymath}
\begin{array}{rcl}
p(\boldsymbol{\beta}|v) & \sim &\mathcal{N}(\bm{\beta}|\mathrm{\mathbf{0}},v^{-1}\mathrm{\mathbf{I}}_{(2M+1)}),
\end{array}
\end{displaymath}where $a_v$ and $b_v$ are hyper-parameters and $v$ plays a similar role as the penalty parameter in SVM.

For simplicity, let $\Theta=(\boldsymbol{\beta},v,\nu_k,\varpi_k,\zeta_k,z_i,\bm{\omega}_i,{\bm{\mu}}_k, {\bm{\Sigma}}_k)$ be the variables of the nonlinear max-margin prediction model. Now, we can formulate our final modal as
\begin{align}
\nonumber
\min\limits_{q(\Omega,\Theta)\in \mathcal{P}} \text{KL}(q(\Omega,\Theta)\|p_0(\Omega,\Theta))-\mathbb{E}_{q(\Omega)}[\text{log}p(\widetilde{\mathrm{\mathbf{X}}}|\Omega)] -\mathbb{E}_{q(\Omega,\Theta)}[\log(\ell(\mathrm{\mathbf{y}}|\mathrm{\mathbf{H}},\Theta)],
\end{align}where $p_0(\Omega,\Theta)$ is the prior, $p_0(\Omega,\Theta)=p_0(\Omega)p_0(\Theta)$ and $p_0(\Theta)$ is the prior of $\Theta$. By solving the optimization above, we get the desired post-data posterior distribution~\cite{ghosh2003bayesian}
\begin{align}
\nonumber q(\Omega,\Theta) = \frac{p_0(\Omega,\Theta)p(\widetilde{\mathrm{\mathbf{X}}}|\Omega)\ell(\mathrm{\mathbf{y}}|\mathrm{\mathbf{H}},\Theta)}{\Xi(\widetilde{\mathrm{\mathbf{X}}},\mathrm{\mathbf{y}})},
\end{align}where $\Xi(\widetilde{\mathrm{\mathbf{X}}},\mathrm{\mathbf{y}})$ is the normalization constant.

\vspace{-0.2cm}
\section{Post-Data Posterior Sampling with HMC}
\vspace{-0.3cm}
As we can see, the post-data posterior above is intractable to compute. Firstly, the pseudo-likelihood function $\ell(\cdot)$ involves a max operater which mixes the posterior inference difficult and inefficient. We introduce the data
augmentation idea~\cite{polson2011data} to solve this problem. Secondly, the form of $\tilde{\varphi}(\mathrm{\mathbf{h}}^n)$ mixes local conjugacy. So we adopt the gradient-based MCMC sampler for faster convergence. Thirdly, we introduce the ditributed DPMM to improve the efficiency. In the following, we devise a hybrid MCMC sampling algorithm that generates a sample from
the post-data posterior distribution of each variable in turn, conditional on the current values of
the other variables. It can be shown that the sequence of samples constitutes a Markov
chain and the stationary distribution of that Markov chain is just the joint posterior.
\vspace{-0.3cm}
\subsection{Updating variables $\bm{\beta}$, \ $\lambda^{n}$, \ $\bm{\omega}_j$ and $\mathrm{\mathbf{h}}^n$}
\vspace{-0.2cm}
In this part, we develop a MCMC sampler for $\bm{\beta}$, $\lambda^{n}$, $\bm{\omega}_j$ and $\mathrm{\mathbf{h}}^n$ by introducing  augmented variables.

\vspace{-0.3cm}
\subsubsection{Data augmentation:}
The pseudo-likelihood function $\ell(\cdot)$ involves a max operater which mixes the posterior inference difficult and inefficient. So we re-express the pseudo-likelihood function into the integration of a function with augmented variable based on the data
augmentation idea:
\begin{align*}
\ell(y^{n}|\tilde{\varphi}(\mathrm{\mathbf{h}}^n),\boldsymbol{\beta}) \! = \!  \int_0^\infty \frac{ \exp\{\frac{-[\lambda^{n}+C(1 - y^{n} \boldsymbol{\beta}^{\mathrm{T}}\tilde{\varphi}(\mathrm{\mathbf{h}}^n))]^2}{2\lambda^{n}}\}
}{\sqrt{2\pi\lambda^{n}}}d\lambda^{n}.\quad \
\end{align*}Then we can get the non-normalized joint distribution of $\mathrm{\mathbf{y}}$ and $\boldsymbol{\lambda}$ conditional on $\mathrm{\mathbf{H}}$ and $\boldsymbol{\Theta}$:
\begin{align*}
\ell(\mathrm{\mathbf{y}},\boldsymbol{\lambda}|\mathrm{\mathbf{H}},\boldsymbol{\Theta}) \!= \! \prod_{n=1}^{N}\frac{ \exp\{\frac{-1}{2\lambda^{n}}[\lambda^{n}+C(1 - y^{n} \boldsymbol{\beta}^{\mathrm{T}}\tilde{\varphi}(\mathrm{\mathbf{h}}^n))]^2\}
}{\sqrt{2\pi\lambda^{n}}}.
\end{align*}

\subsubsection{\textbf{{Sampling $\bm{\beta}$ :}}}
The conditional distribution of $\bm{\beta}$ is
\begin{align}
\label{sample-beta}
\nonumber q(\boldsymbol{\beta}|v,\mathrm{\mathbf{H}},\bm{\omega}, \mathrm{\mathbf{y}},\bm{\lambda})  & \sim  p(\boldsymbol{\beta}|v) \prod_{n=1}^{N} \ell(y^{n},\lambda^{n}|\tilde{\varphi}(\mathrm{\mathbf{h}}^n),\boldsymbol{\beta})\\
& \propto  \exp(-\frac{v||\bm{\beta}||^2}{2}-\sum_{n=1}^{N}\frac{[\lambda^{n}+\Lambda]^2}{2\lambda^{n}})
,\end{align}where $\Lambda=C(1 - y^{n} \boldsymbol{\beta}^{\mathrm{T}}\tilde{\varphi}(\mathrm{\mathbf{h}}^n))$. This conditional distribution is a Gaussian distribution with covariance  $\bm{\Sigma}_{\bm{\beta}}=\{v\mathrm{\mathbf{I}}_{2M+1}+\sum_{n=1}^{N}\frac{C^2{\tilde{\varphi}(\mathrm{\mathbf{h}}^n}){\tilde{\varphi}(\mathrm{\mathbf{h}}^n)}^{\mathrm{T}}}{\lambda^{n}}\}^{-1}$ and mean ${\bm{\mu}}_{\bm{\beta}}=\bm{\Sigma}_{\bm{\beta}}\sum_{n=1}^{N}(\frac{C{\lambda^n}+C^2}{\lambda^{n}})y^n{\tilde{\varphi}(\mathrm{\mathbf{h}}^n))}$.

\subsubsection{\textbf{{Sampling $\lambda^{n}$ :}}}
The conditional distribution over the augmented variable $\lambda^{n}$ is a generalized inverse Gaussian distribution:
\begin{align}
\label{sample-lambda}
\nonumber q(\lambda^n|\mathrm{\mathbf{h}}^n,\bm{\omega},y^n,\bm{\beta}) & \propto   \exp(-\frac{1}{{2\lambda^{n}}}{\{\lambda^{n}+C[1 - y^{n} \boldsymbol{\beta}^{\mathrm{T}}\tilde{\varphi}(\mathrm{\mathbf{h}}^n)]\}^2})
\\
&  \sim  \mathcal{GIG}(\lambda^{n}|\frac{1}{2},1,C^2[1 - y^{n} \boldsymbol{\beta}^{\mathrm{T}}\tilde{\varphi}(\mathrm{\mathbf{h}}^n)]^2).
\end{align}

\subsubsection{\textbf{Sampling $\bm{\omega}_j$, \ $\mathrm{\mathbf{h}}^n$:}}
Unfortunately, we find the conditional distribution over $\bm{\omega}_j$
\begin{align}
\label{sampe-omega}
\nonumber
&q(\bm{\omega}_j|\mathrm{\mathbf{H}},\bm{\lambda},\bm{\beta},\mathrm{\mathbf{y}}) \sim p(\bm{\omega}_j|\bm{\mu}_{z_j},\bm{\Sigma}_{z_j})\prod_{n=1}^{N} \ell(y^{n},\lambda^{n}|\tilde{\varphi}(\mathrm{\mathbf{h}}^n),\boldsymbol{\beta})\\
& \
\propto \exp \{ -\frac{(\bm{\omega}_j-\bm{\mu}_{z_j})^{T}{\bm{\Sigma}^{-1}_{z_j}}(\bm{\omega}_j-\bm{\mu}_{z_j})}{2}-\sum_{n=1}^{N}\frac{[\lambda^{n}+\Lambda]^2}{2\lambda^{n}}\}
,\end{align}
where $\tilde{\varphi}(\mathrm{\mathbf{h}}^n)=({{\varphi}(\mathrm{\mathbf{h}}^n)}^{\mathrm{ T }},1)^{\mathrm{ T }}=\{ \frac{1}{\sqrt{M}}[\cos(\bm{\omega}_1^{\mathrm{ T }}\mathrm{\mathbf{h}}^n),\cdots,$ $\cos(\bm{\omega}_M^{\mathrm{ T }}\mathrm{\mathbf{h}}^n),\sin(\bm{\omega}_1^{\mathrm{ T }}\mathrm{\mathbf{h}}^n),\cdots,$ $\sin(\bm{\omega}_M^{\mathrm{ T }}\mathrm{\mathbf{h}}^n)],1\}^{\mathrm{ T }}$ is too complex and the prior is non-conjugated. So it is hard to get the analytical
form of the above distribution. To generate a sample from $q(\bm{\omega}_j|\mathrm{\mathbf{H}},\bm{\lambda},\bm{\beta},\mathrm{\mathbf{y}}) $ with its non-normalized density, we appeal to the HMC method~\cite{neal2011mcmc}. Sampling from a distribution with HMC requires translating the density function for this distribution to a potential energy function and introducing 'momentum' variables to go with the original variables of interest ('position' variables). We need to constitute a Markov chain. Each iteration has two steps. In the first step, we sample the momentum which needs the gradient of the potential
energy function. In the second step, we do a Metropolis update with a proposal found by using Hamiltonian dynamics.

Similar to $\bm{\omega}_j$, we also find the conditional posterior distribution over $\mathrm{\mathbf{h}}^n$
\begin{align}
\label{sample-h}
\nonumber
& q(\mathrm{\mathbf{h}}^n|\bm{\omega},\bm{\lambda},\bm{\beta},\mathrm{\mathbf{y}},\mathrm{\mathbf{X}}) \sim p(\mathrm{\mathbf{h}}^n|\mathrm{\mathbf{0}},\mathrm{\mathbf{I}})\prod_{i=1}^{N_v}p(\mathrm{\mathbf{x}}^n_i
|\mathrm{\mathbf{W}}_i,\mathrm{\mathbf{h}}^n,\mathrm{\mathbf{u}}^n_i,\tau_i) \ell(y^{n}|\tilde{\varphi}(\mathrm{\mathbf{h}}^n),\boldsymbol{\beta})\\
&\ \ \ \ \ \ \ \ \ \ \ \ \
\propto \exp\{ -\frac{||\mathrm{\mathbf{h}}^n||^{2}}{2} -\sum_{i=1}^{N_v}\frac{{\tau_i}||\mathrm{\mathbf{x}}_i^n-\mathrm{\mathbf{W}}_i\mathrm{\mathbf{h}}^n-\mathrm{\mathbf{V}}_i\mathrm{\mathbf{u}}_i^n||^2}{2} - \frac{[\lambda^{n}+\Lambda]^2}{2\lambda^{n}} \}
,\end{align} doesn't have the analytical form. So we can sample $\mathrm{\mathbf{h}}^n$ with the HMC sampler.

\subsection{Updating the variables in distributed DPMM}
Unlike a Gibbs sampler using the marginal representation of the DP mixtures, slice sampling methods \cite{walker2007sampling,kalli2011slice,ge2015distributed} employ the random measure $G$ directly. It consists of the imputation of $G$ and subsequent Gibbs sampling of the component assignments from their posteriors.
To be able to represent the infinite number of components in $G_0$, we have to introduce some auxiliary (slice) variables $t_j$, $j=1,\cdots,M$. Through introducing an auxiliary variable $t_j$, the joint density
of $\bm{\omega}_j$ and the latent variable $t_j$ becomes
\vspace{-0.2cm}
\begin{align*}
p(\bm{\omega}_j,t_j|\bm{\varpi},\bm{\zeta})=\sum_{k=1}^{\infty}{{\varpi}_k}
\text{Unif}(t_j|0,{{\varpi}_k})p(\bm{\omega}_j|{\zeta}_k)
=\sum_{k=1}^{\infty}\mathds{1}(t_j\leq{\varpi}_k)p(\bm{\omega}_j|{\zeta}_k),
\end{align*}where $\mathds{1}(\cdot)$ is the indicator function. We can easily verify that when $t_j$ is integrated over, the joint density is equivalent to $p(\bm{\omega}_j|\bm{\varpi},\bm{\zeta})$. Thus the interesting
fact is that given the latent variable $t_j$, the number of mixtures
needed to be represented is finite.

Through introducing the slice variable, the conditional posterior distribution of $\bm{\omega}_j$ becomes independent so it is possible to derive a parallel sampler for the DP mixture model under the Map-Reduce framework. The latent variable $t_j$ does not
change the marginal distribution of other variables, thus the
sampler target the correct posterior distribution. Another
important feature of the slice sampler is that it enables direct
inference of random measure $G$.

\emph{\textbf{Step 1:}} Sample slice variables and find the minimum
\begin{align}
\label{DPMM-1}
t_j\sim\text{Unif}(0,\varpi_{z_j}),\quad  \forall\ j=1,...,M; \quad\quad     t^*=\min_j t_j.
\end{align}

\emph{\textbf{Step 2:}} Create new components through stick breaking until $\varpi^* < t^*$ with $\varpi^*$  being the remaining stick length and $K^*$ the number of instantiated components
\begin{align}
\label{DPMM-2}
\nonumber K^* \leftarrow K^* +1,\quad  \nu_{K^*}\sim \text{Beta}(1,\alpha), \quad\quad\quad\quad \\
\varpi_{K^*} = \varpi^*\nu_{K^*},\quad  \zeta_{K^*}\sim G_0, \quad   \varpi^* \leftarrow \varpi^*(1-\nu_{K^*}).
\end{align}

\emph{\textbf{Step 3:}} Sample the component assignment $z_j$
\begin{align}
\label{DPMM-3}
p(z_j=k|\bm{\omega}_j,t_j,\bm{\varpi},\bm{\zeta}) \propto  \left\{\begin{array}{@{}l@{\ \ \ }l} p(\bm{\omega}_j|\zeta_k)  & \quad  \text{if} \  \varpi_{k}\geq t_j \\ 0 & \quad \text{otherwise}.  \end{array}
\right .
\end{align}

\emph{\textbf{Step 4:}} For each active component $k$, sample component parameters $\zeta_k$
\begin{align}
\label{DPMM-4}
\zeta_k|\mathrm{\mathbf{z}},G_0 \sim  G_0(\zeta_k)\prod_{j:z_j=k} p(\bm{\omega}_j|\zeta_k).
\end{align}

\emph{\textbf{Step 5:}} For each component $k$, sample component weights:
\begin{align}
\label{DPMM-5}
\bm{\varpi}|\mathrm{\mathbf{z}},\alpha \sim  \text{Dir}(s_1,s_2,...,s_K,\alpha).
\end{align}
where $s_k$ is the number of variables $\bm{\omega}_j$ assigned to component $k$, and $K$ is the number of active components. Our parallel sampler in distributed DPMM is similar as that in ~\cite{ge2015distributed}, thus is omitted here.

\vspace{-0.3cm}
\subsection{Updating the other variables}
In this part, the other variables all have the conjugate prior. So we get the analytic conditional posterior distributions of them.

\subsubsection{\textbf{Sampling $\mathrm{\mathbf{u}}_i^n$ :}}
The conditional posterior of $\mathrm{\mathbf{u}}_i^n$ is
\begin{align}
\label{sample-u}
\nonumber
q({\mathrm{\mathbf{u}}_i^n}|\tau_i,\mathrm{\mathbf{h}}^n,\mathrm{\mathbf{x}}^n_i,\mathrm{\mathbf{W}}_i) &\sim p(\mathrm{\mathbf{u}}_i^n|\mathrm{\mathbf{0}},\mathrm{\mathbf{I}})p(\mathrm{\mathbf{x}}^n_i
|\mathrm{\mathbf{W}}_i,\mathrm{\mathbf{h}}^n,\mathrm{\mathbf{u}}^n_i,{\mathrm{\mathbf{V}}_i},\tau_i)\\
&\propto \exp\{ -\frac{1}{2}({||{\mathrm{\mathbf{u}}_i^n}||^2}-{{\tau_i}||\mathrm{\mathbf{x}}_i^n-\mathrm{\mathbf{W}}_i\mathrm{\mathbf{h}}^n-\mathrm{\mathbf{V}}_i\mathrm{\mathbf{u}}_i^n||^2}) \},
\end{align} a Gaussian distribution with covariance ${\bm{\Sigma}}_{\mathrm{\mathbf{u}}_i^n}=(\mathrm{\mathbf{I}}+{\tau_i}{\mathrm{\mathbf{V}}_i}^{\mathrm{T}}{\mathrm{\mathbf{V}}_i})^{-1}$ and mean $ {\bm{\mu}}_{\mathrm{\mathbf{u}}_i^n}={\bm{\Sigma}}_{\mathrm{\mathbf{u}}_i^n}[{\mathrm{\mathbf{V}}_i}^{\mathrm{T}}(\mathrm{\mathbf{x}}_i^n-\mathrm{\mathbf{W}}_i\mathrm{\mathbf{h}}^n){\tau_i}]$.

\subsubsection{\textbf{Sampling $\mathrm{\mathbf{W}}_i$:}}
The conditional posterior distribution of $\mathrm{\mathbf{W}}_i$ is proportional to the prior times the likelihood:
\begin{align}
\label{sample-w}
\nonumber & q({\mathrm{\mathbf{w}}_{i,\cdot j}}|r_{i,j},\tau_i,\mathrm{\mathbf{H}},\mathrm{\mathbf{U}}_i,\mathrm{\mathbf{X}}_i,{\mathrm{\mathbf{V}}_i}) \sim p({\mathrm{\mathbf{w}}_{i,\cdot j}}|\mathrm{\mathbf{0}},{r_{i,j}^{-1}}\mathrm{\mathbf{I}})p(\mathrm{\mathbf{X}}_i
|\mathrm{\mathbf{W}}_i,\mathrm{\mathbf{H}},\mathrm{\mathbf{U}}_i,{\mathrm{\mathbf{V}}_i},\tau_i)\\
& \ \ \ \ \ \ \ \ \ \ \ \ \ \ \ \ \ \ \ \ \ \ \ \
\propto  \exp\{ -\frac{1}{2}({{r_{ij}}||{\mathrm{\mathbf{w}}_{i,\cdot j}}||^2+{ \sum\limits_{n=1}^{N}{\tau_i} ||\mathrm{\mathbf{x}}_i^n-\mathrm{\mathbf{W}}_i\mathrm{\mathbf{h}}^n-\mathrm{\mathbf{V}}_i\mathrm{\mathbf{u}}_i^n||^2}} )\},
\end{align}where $j=1,\cdots,m$. This is a Gaussian distribution with covariance
${\bm{\Sigma}}_{\mathrm{\mathbf{w}}_{i,\cdot j}} = (r_{ij}+{\tau_i}||{\mathrm{\mathbf{h}}_{j \cdot}}||^2)^{-1}\mathrm{\mathbf{I}}$ and mean
${\bm{\mu}}_{\mathrm{\mathbf{w}}_{i,\cdot j}} =  {\bm{\Sigma}}_{\mathrm{\mathbf{w}}_{i,\cdot j}}
 \{\sum\limits_{n=1}^N [\mathrm{\mathbf{x}}_i^n-{\sum\limits_{k\neq j}^m}{\mathrm{\mathbf{w}}_{i,\cdot k}}{h_{kn}}-{\mathrm{\mathbf{V}}}_i{{u_{i,jn}}}] {\tau_i}{h_{jn}}\}$.

\subsubsection{\textbf{Sampling $\mathrm{\mathbf{V}}_i$ :}}
Similar to $\mathrm{\mathbf{W}}_i$, conditional posterior of $\mathrm{\mathbf{V}}_i$ is a Gaussian distribution
\begin{align}
\label{sample-v}
\nonumber
& q({\mathrm{\mathbf{v}}_{i,\cdot j}}|\eta,\tau_i,\mathrm{\mathbf{H}},\mathrm{\mathbf{U}}_i,\mathrm{\mathbf{X}}_i,{\mathrm{\mathbf{W}}_i}) \sim p({\mathrm{\mathbf{v}}_{i,\cdot j}}|\mathrm{\mathbf{0}},{{\eta}^{-1}}\mathrm{\mathbf{I}})p(\mathrm{\mathbf{X}}_i
|\mathrm{\mathbf{W}}_i,\mathrm{\mathbf{H}},\mathrm{\mathbf{U}}_i,{\mathrm{\mathbf{V}}_i},\tau_i)\\
& \ \ \ \ \ \  \ \ \ \ \ \ \ \ \ \ \ \ \ \ \ \ \
 \propto  \exp\{ -\frac{1}{2}({{\eta}||{\mathrm{\mathbf{v}}_{i,\cdot j}}||^2+{ \sum\limits_{n=1}^{N}{\tau_i} ||\mathrm{\mathbf{x}}_i^n-\mathrm{\mathbf{W}}_i\mathrm{\mathbf{h}}^n-\mathrm{\mathbf{V}}_i\mathrm{\mathbf{u}}_i^n||^2}} )\},
\end{align}with covariance ${\bm{\Sigma}}_{\mathrm{\mathbf{v}}_{i,\cdot j}}=(\eta+{\tau_i}||{\mathrm{\mathbf{u}}_{i,\cdot j}}||^2)^{-1}\mathrm{\mathbf{I}}$ and mean ${\bm{\mu}}_{\mathrm{\mathbf{v}}_{i,\cdot j}}={\bm{\Sigma}}_{\mathrm{\mathbf{v}}_{i,\cdot j}}
 \{\sum_{n=1}^N [\mathrm{\mathbf{x}}_i^n-{\sum\limits_{k\neq j}^m}{\mathrm{\mathbf{v}}_{i,\cdot k}}{u_{i,kn}}-{\mathrm{\mathbf{V}}}_i{{u_{i,jn}}}] {\tau_i}{h}_{jn}\}$.

\subsubsection{\textbf{Sampling $\mathrm{\mathbf{r}}_i$:}}
For each ${r}_{ij},j=1,\cdots,m$, its conditional distribution is
\begin{align}
\label{sample-r}
\nonumber
q(r_{ij}) & \sim p(r_{ij}|a_r,b_r)p({\mathrm{\mathbf{w}}}_{i,\cdot j}|{\mathrm{\mathbf{0}}},r_{ij}^{-1}{\mathrm{\mathbf{I}}})
\\
& \propto r_{ij}^{a_r-1+\frac{D_i}{2}}\exp\{-r_{ij}(b_r+\sum\limits_{d=1}^{D_i}\frac{1}{2}{||{\mathrm{\mathbf{w}}}_{i,\cdot j}||^2)}\},
\end{align} a Gamma distribution with the shape and rate parameter $a_{r_{ij}}=a_r+\frac{D_i}{2}, \ b_{r_{ij}}=b_r+\sum\limits_{d=1}^{D_i}\frac{1}{2}{||{\mathrm{\mathbf{w}}}_{i, \cdot j}||^2}$.

\subsubsection{\textbf{Sampling $\tau_i$ :}}
Similar to sampling $\mathrm{\mathbf{r}}_i$, the conditional distribution of $\tau_i$ is a Gamma distribution
\begin{align}
\label{sample-tau}
\nonumber
q(\tau_i) & \sim p(\tau_i|a_\tau,b_\tau)p(\mathrm{\mathbf{X}}_i
|\mathrm{\mathbf{W}}_i,\mathrm{\mathbf{H}},\mathrm{\mathbf{U}}_i,{\mathrm{\mathbf{V}}_i},\tau_i)
\\
& \propto {\tau_i}^{a_{\tau}-1+\frac{ND_i}{2}}\exp\{-{\tau}(b_{\tau}+{ \sum\limits_{n=1}^{N} \frac{1}{2} ||\mathrm{\mathbf{x}}_i^n-\mathrm{\mathbf{W}}_i\mathrm{\mathbf{h}}^n-\mathrm{\mathbf{V}}_i\mathrm{\mathbf{u}}_i^n||^2)}\},
\end{align}
with the shape and rate parameter $a_{\tau_i}=a_{\tau}+\frac{ND_i}{2}, \ b_{\tau_i}=b_{\tau}+{ \sum\limits_{n=1}^{N}\frac{1}{2} ||\mathrm{\mathbf{x}}_i^n-\mathrm{\mathbf{W}}_i\mathrm{\mathbf{h}}^n-\mathrm{\mathbf{V}}_i\mathrm{\mathbf{u}}_i^n||^2}$.

We summarize the above post-data posterior sampling process in Alg. \ref{alg:M3LAK}.
\vspace{-0.3cm}
\renewcommand{\algorithmicrequire}{\textbf{Input:}}
\renewcommand{\algorithmicensure}{\textbf{Output:}}
\begin{algorithm}
\caption{Post-Data Posterior Sampling  for {M$^3$LAK}}
\label{alg:M3LAK}
\begin{algorithmic}[1]
\REQUIRE ~~ \\
the multi-view data $\{{\mathrm{\mathbf{X}}}_i\}_{i=1}^{N_v}$, the label vector $\mathrm{\mathbf{y}}$, the subspace dimension $m$, the number of random features $M$, the hyper-parameters $\eta$, $\alpha$, \ $a_r$, \ $b_r$, \ $a_{\tau}$,
 $b_{\tau}$, \ $v$, and the maximal number of iterations $maxIter$.
\\
\ENSURE ~~\\ 
$\{\tau_i, \mathrm{\mathbf{W}}_i, \mathrm{\mathbf{V}}_i\}_{i=1}^{N_v}$, \ $\{\bm{\omega}_j\}_{j=1}^{M}$, \ $\bm{\beta}$ \\
\renewcommand{\algorithmicensure}{\textbf{Method:}}
\ENSURE ~~\\ 
\STATE Initialize all variables $\bm{\beta}$, $\bm{\lambda}$, $\{\bm{\omega}_j, z_j\}_{j=1}^{M}$, $\bm{\zeta}$, $\bm{\varpi}$, $\bm{\nu}$, $\mathrm{\mathbf{U}}$, $\mathrm{\mathbf{W}}$, $\mathrm{\mathbf{V}}$, $\{\mathrm{\mathbf{r}}_i, \tau_i\}_{i=1}^m$;

\FOR {$iter=1$ to $maxIter$}

\STATE Update $\bm{\beta}$ according to Eq.\eqref{sample-beta};\\

\STATE Update $\bm{\lambda}$ according to Eq.\eqref{sample-lambda};\\

\STATE Update $\{\bm{\omega}_j\}_{j=1}^{M}$, \ $\mathrm{\mathbf{H}}$  using the HMC sampler in ~\cite{neal2011mcmc};\\

\STATE Update $( \ \{z_j\}_{j=1}^{M}, \ \bm{\zeta}, \  \bm{\varpi},
\ \bm{\nu} \ )$ through introducing some auxiliary variables $\{t_j\}_{j=1}^{M}$ according to Eq.\eqref{DPMM-1}-Eq.\eqref{DPMM-5}, which can be paralleled in Map-Reduce framework similar as that in~\cite{ge2015distributed}; \\

\STATE Update $\mathrm{\mathbf{U}}$ according to Eq.\eqref{sample-u};\\

\STATE Update $\mathrm{\mathbf{W}}$ according to Eq.\eqref{sample-w};\\

\STATE Update $\mathrm{\mathbf{V}}$ according to Eq.\eqref{sample-v};\\
\STATE Update \ $\{\mathrm{\mathbf{r}}_i\}_{i=1}^m$ \ according to Eq.\eqref{sample-r};\\

\STATE Update \ $\{\tau_i\}_{i=1}^m$ \  according to Eq.\eqref{sample-tau};\\
\ENDFOR

\RETURN $\{\tau_i, \mathrm{\mathbf{W}}_i, \mathrm{\mathbf{V}}_i\}_{i=1}^{N_v}$, \ $\{\bm{\omega}_j\}_{j=1}^{M}$, \ $\bm{\beta}$ . 
\end{algorithmic}
\end{algorithm}

\vspace{-0.5cm}
\subsubsection{\textbf{Computational Complexity:}}
In our post-data posterior sampling, the dominant computation is spent on sampling latent shared variables $\mathrm{\mathbf{H}}$. In each round of parameter sampling, our algorithm consumes $O(NL\sum\limits_{i=1}^{N_v}D_i(m+K_i))$ operations where $L$ is the number of steps for the leapfrog method in HMC. So the computational complexity of our algorithm M$^3$LAK is linear w.r.t. the number of instances $N$. The source code of this work is available on \footnote[1]{https://github.com/hezi73/M3LAK}.

\vspace{-0.1cm}
\section{Experiments}
\vspace{-0.1cm}
In this section, we evaluate our proposed model (M$^3$LAK) on various classification
tasks. The code for M$^3$LAK was written purely in Matlab and all experiments were performed on a desktop with 2.10 GHz CPU and 128 GB memory.

\vspace{-0.2cm}
\subsection{Data Description}
\vspace{-0.1cm}
The Flickr dataset  contains 3,411 images of 13 animals~\cite{chen2012large}. For each image, two types of features are extracted, including 634-dim real-valued features and 500-dim bag of word SIFT features.
The NUS-WIDE dataset is a subset selected from~\cite{nus-wide-civr09}. NUS-WIDE dataset contains 21935 web images that belongs to three categories (`water', `vehicle', `flowers'). Each image includes six types of low-level features (64-D color histogram, 144-D color correlogram, 73-D edge direction histogram, 128-D wavelet texture, 225-D block-wise color moments).
Trecvid contains 1,078 manually labeled video shots that belongs to five categories~\cite{chen2012large}. And each shot is represented by a 1,894-dim binary vector of text features and a 165-dim vector of HSV color histogram. The web-page data set has two views, including the content features of the web pages and the link features exploited from the link structures. This data set consists of web pages from computer science department in three universities, i.e., Cornell, Washington, Wisconsin.

We transform these multi-class data sets into binary ones by following the way in ~\cite{Zhuang2012Multi}. For example, the web-page classification with five categories (`course', `faculty', `student', `project', `staff'), we select category `student' as a group and the other four categories as another group, since the number of examples in category `student' is similar with the one belonging to the other four categories. The other data sets are similarly constructed. The details of these data sets are shown in Table 1.

\begin{table}[h]
\caption{Detail description of datasets.}
\centering
\renewcommand{\arraystretch}{1}
\setlength\tabcolsep{7pt}
\begin{tabular}{c c c c c c c}
\hline
Datasets&Trecvid&Flickr&Cornell&Washington&Wisconsin&NUS-WIDE\\
\hline
\#Size&1078&3411&195&217&262&21935\\
\#D$_1$&1894&634&1703&1703&1703&64\\
\#D$_2$&165&500&195&217&262&144\\
\#D$_3$&-&-&-&-&-&73\\
\#D$_4$&-&-&-&-&-&128\\
\#D$_5$&-&-&-&-&-&225\\
\hline
\end{tabular}
\end{table}
\vspace{-0.5cm}

\subsection{Baselines}
\vspace{-0.1cm}
We compare our model with five competitors:
\vspace{-0.1cm}
\begin{itemize}
\item BM$^2$SMVL~\cite{heonline}:
a linear Bayesian max-margin subspace multi-view learning method.

\item MVMED~\cite{sun2013multi}:
a multi-view maximum entropy discrimination model.

\item VMRML~\cite{quang2013unifying}:
a vector-valued manifold regularization multi-view learning method.

\item MMH~\cite{chen2012large}:
a predictive latent subspace Markov network multi-view learning model.

\item BEMKL~\cite{Gonen2012Bayesian}:
a state-of-the-art Bayesian multiple kernel learning method which we use for multi-view learning. For each view, we construct Gaussian kernels with 21 different widths $\{ {2^{-10}, 2^{-9}, \cdots , 2^{10}} \}$ on all features by using its public implementation \footnote[2]{http://users.ics.aalto.fi/gonen/bemkl/}.
\end{itemize}

\vspace{-0.1cm}
\subsection{Experimental Setting}
\vspace{-0.1cm}
In M$^3$LAK, we perform 5-fold cross-validation on training set to decide the regularization parameter $C$ from the integer set $\{1,\cdots, 10\}$ for each data set. The rest parameters are fixed as follows for all data sets, i.e., $m=$ 20, $M=$ 100, $\eta=$ 1e+3, $\alpha=$ 1, $a_r=$ 1e-1, $a_\tau=v=$ 1e-2, $b_\tau=b_r=$ 1e-5. For fair comparison, the subspace dimension $m$ in BM$^2$SMVL and MMH is also fixed to $20$ for all data sets. For BM$^2$SMVL, the regularization parameter `$C$' is chosen from the integer set $\{1,2,3\}$ by 5-fold cross-validation on training set according to its original paper. For MVMED, we choose `$c$' from $2^{[-5:5]}$ for each data set as suggested in its original paper. For VMRML, the parameters are set as the default values in its paper. For RBF kernel's parameter of MVMED and VMRML, we carefully tune them on each data set separately. For MMH, we tune its parameters as suggested in its original papers.

On each data set except NUS-WIDE, we conduct 10-fold cross validation for all the algorithms, where nine folds of the data are used for training while the rest for testing. And we run 1000 MCMC iterations of M$^3$LAK and use the samples collected in the last 200 iterations for prediction.
The averaged accuracies over these 10 runs are reported in Table 2. We use the same training/testing split of the NUS-WIDE data set as in~\cite{nus-wide-civr09}. To compare the time efficiency, we conduct experiments on NUS-WIDE with different numbers of the training data (1000, 2000, 3000, 5000, 10000) and keep the testing data the same. The training data are selected randomly from the training split of NUS-WIDE. Because the data size of NUS-WIDE is big, we run 200 MCMC iterations of M$^3$LAK and use the samples collected in the last 100 iterations for prediction. The result is reported in Table 3 in terms of testing accuracy and training time. Since MMH can only deal with two-view data in its code\footnote[3]{http://bigml.cs.tsinghua.edu.cn/~ningchen/MMH.htm}, its result is missing for NUS-WIDE in Table 3.

\vspace{-0.1cm}
\begin{table*}[!htb]
\small
\caption{ Comparison of test accuracies (mean $\pm$ std) on all datasets. Bold face indicates highest accuracy.}
\centering
\renewcommand{\arraystretch}{1.5}
\begin{tabular}{ccccccc}
\hline
&MMH&MVMED&VMRML&BM$^{2}$SMVL&BEMKL&M$^3$LAK\\
\hline
Trecvid-a&{ .939 $\pm$ .066 }&{ .913 $\pm$ .030 }&{ .921 $\pm$ .028 }&{ .901 $\pm$ .019 }&{ .944 $\pm$ .033 }&{ \textbf{.954 $\pm$ .033} }\\
Trecvid-b&{ \textbf{.944 $\pm$ .034} }&{ .920 $\pm$ .068 }&{ .932 $\pm$ .063 }&{ .912 $\pm$ .059 }&{ .932 $\pm$ .060 }&{  .940 $\pm$ .054}\\
Flickr-a&{ .820 $\pm$ .085 }&{ .854 $\pm$ .088 }&{ .800 $\pm$ .068 }&{ .827 $\pm$ .070 }&{ \textbf{.857 $\pm$ .087} }&{  .855 $\pm$ .056}\\
Flickr-b&{ .823 $\pm$ .055 }&{ .858 $\pm$ .065 }&{ .834 $\pm$ .044 }&{ .856 $\pm$ .048 }&{ \textbf{.871 $\pm$ .046} }&{  \textbf{.871 $\pm$ .034}}\\
Cornell&{  .862 $\pm$ .063 }&{ .861 $\pm$ .080 }&{ .882 $\pm$ .069 }&{ .882 $\pm$ .072 }&{ .861 $\pm$ .060 }&{ \textbf{.903 $\pm$ .065} }\\
Washington&{ .909 $\pm$ .042 }&{ .852 $\pm$ .077 }&{ .874 $\pm$ .069 }&{ .896 $\pm$ .055 }&{ .874 $\pm$ .052 }&{ \textbf{.913 $\pm$ .058} }\\
Wisconsin&{ .906 $\pm$ .043 }&{ .872 $\pm$ .047 }&{ .883 $\pm$ .073 }&{ .921 $\pm$ .072 }&{ .902 $\pm$ .072 }&{ \textbf{.936 $\pm$ .050} }\\
Average&{ .886  }&{ .876  }&{ .875 }&{ .885 }&{ .892 }&{ \textbf{.910 } }\\
\hline
\end{tabular}
\end{table*}
\vspace{-0.5cm}

\begin{table*}[!th]
\small
\centering
\renewcommand{\arraystretch}{1}
\setlength\tabcolsep{0.6pt}
\caption{Comparison of various multi-view methods on binary classification tasks. Each element in the table shows the testing accuracies/training times on the NUS-WIDE dataset. `N/A' means that no result returns after 24 hours. `-' means out of memory. All experiments were conducted in Matlab.}
\label{benchmark classification}
\newcommand{\minitab}[2][l]{\begin{tabular}{#1}#2\end{tabular}}
\begin{tabular}{ccccccccc}
\hline
                            \multirow{2}*{Algorithm}  & \multirow{2}*{{Metric}}
                           & N$_{train}$=1000  & N$_{train}$=2000   & N$_{train}$=3000
                           & N$_{train}$=5000     & N$_{train}$=10000   \\
                         & & N$_{test}$=8033 & N$_{test}$=8033 & N$_{test}$=8033
                           & N$_{test}$=8033 & N$_{test}$=8033 \\
\hline
\multirow{2}*{BM$^2$SMVL}
& \ Test-Acc (\%)       &72.02          & 77.12         &77.54          &77.29          &77.67          \\
& \ Train-Time (s)\      &{234}       &{277}         &{340}     &{490}          &\textbf{776}\\
\arrayrulecolor{mygray} \hline \arrayrulecolor{black}
\multirow{2}*{\minitab[c]{ VMRML}}
& \ Test-Acc (\%)\       &74.58          & 76.09         &76.29          &76.43          &76.68          \\
& \ Train-Time (s)\      &\textbf{4}       &\textbf{18}  &\textbf{50}       &\textbf{210}       &1335 \\
\arrayrulecolor{mygray} \hline \arrayrulecolor{black}
\noalign{\vskip 0.03in}
\multirow{2}*{\minitab[c]{ MVMED }}
& \ Test-Acc (\%)\       &67.09          &76.66          &N/A           &N/A           &N/A \\
& \ Train-Time (s)\      &1363      &9721       &N/A       &N/A       &N/A\\
\arrayrulecolor{mygray} \hline \arrayrulecolor{black}
\multirow{2}*{\minitab[c]{ BEMKL }}
& \ Test-Acc (\%)      &-          &-           &-      &-          &-          \\
& \ Train-Time (s)\      &-        &-    &-       &-        &-  \\
\arrayrulecolor{mygray} \hline \arrayrulecolor{black}
\multirow{2}*{\minitab[c]{ M$^3$LAK }}
& \ Test-Acc (\%)\       &\textbf{75.46}          &\textbf{77.22}          &\textbf{77.47}          &\textbf{78.08}          &\textbf{78.68}           \\
& \ Train-Time (s)\      &402          &844          &1283 &2211   &5541 \\
\arrayrulecolor{mygray} \hline \arrayrulecolor{black}
\hline
\end{tabular}
\end{table*}

\subsection{Experimental Results}
\vspace{-0.1cm}
We have the following insightful observations:
\begin{itemize}
\item[-] M$^3$LAK consistently outperforms BM$^2$SMVL. The reason may be that BM$^2$SMVL is a linear multi-view method with limited modeling capabilities.
\item[-] On most data sets, M$^3$LAK performs better than MMH. This may be because that unlike MMH which are under the maximum entropy discrimination framework, and can not infer the penalty parameter of max-margin models in a Bayesian style, our method is based on the data augmentation idea for max-margin learning, which allows us to automatically infer the weight parameters and the penalty parameter.
\item[-] M$^3$LAK has better performance than the single kernel multi-view learning methods VMRML and MVMED on all considered data sets. The reason may be that M$^3$LAK infers a posterior under the Bayesian framework instead of a point estimate as in VMRML. With Baysian model averaging over the posterior, we can make more robust predictions than VMRML. And MVMED is also under the maximum entropy discrimination framework, and can not infer the penalty parameter of max-margin models in a Bayesian style.
\item[-] M$^3$LAK performs better than BEMKL on most data sets. BEMKL's performance may be limited by its mean-field assumption on the approximate posterior and the absence of max-margin principle while M$^3$LAK introduces the popular max-margin principle which has a great generalization ability. Although BEMKL performs better than M$^3$LAK on some data sets, BEMKL scales cubically with the number of kernels and scales cubically with the number of training samples. Besides, BEMKL needs to store many Gram matrix to get good performances. However, storing too many Gram matrix leads to out of memory on commonly used computers. For example, it needs to calculate $21\times 5 = 105$ Gram matrixes on NUS-WIDE not only for the training data but also for the large amount of testing data. So BEMKL leads to out of memory on NUS-WIDE dataset.
\item[-]  As shown in Table 3, M$^3$LAK scales linearly with the number of training data $N$ which coincides with the computational complexity of M$^3$LAK discussed in Section Computational Complexity. The linear multi-view method BM$^2$SMVL also scales linearly with $N$, but it performs worse than M$^3$LAK on all considered data. Although the training time efficiency of VMRML is better than that of M$^3$LAK in Table 3, it seems that VMRML scales squarely with the number of training data and VMRML needs to store 5 Gram matrixes for both training and testing data on NUS-WIDE data set. Further more, we conduct a experiment with 20000 training data, VMRML is out of memory while M$^3$LAK still works. Besides, M$^3$LAK performs better than VMRML on all considered data.
\end{itemize}

\vspace{-0.5cm}
\subsection{Parameter Study and Convergence}
\vspace{-0.1cm}
We study the performance change of the three subspace learning methods (BM$^2$SMVL, MMH and M$^3$LAK). Performances change when the subspace dimension $m$ varies on two datasets (Cornell and Wisconsin). The averaged results are shown in Figure 1. As we can see, different methods prefer  different values of $m$. On some data sets, when $m$ becomes too large, the performances of these three methods become poor. When $m$ ranges from 5 to 30, M$^3$LAK performs better than other subspace learning methods in general.

Also, we set $m$ as 20 and study the influence of regularization parameter $C$. From the results in Figure 2 (a) , we can find that different data sets may prefer different values of $C$. $C$ balances the nonlinear classifier with adaptive kernel and the multi-view latent variable model, so M$^3$LAK cannot get the best performance when $C$ is too large or small.

Figure 2 (b) shows the convergency of M$^3$LAK on two data sets. We find that M$^3$LAK has a fast convergence rate, which we contribute to the efficient gradient-based HMC sampler~\cite{neal2011mcmc}.
\vspace{-0.1cm}
\begin{figure}[!htbp]
\centering
\subfigure[Cornell] {\includegraphics[height=1.7in,width=2in]{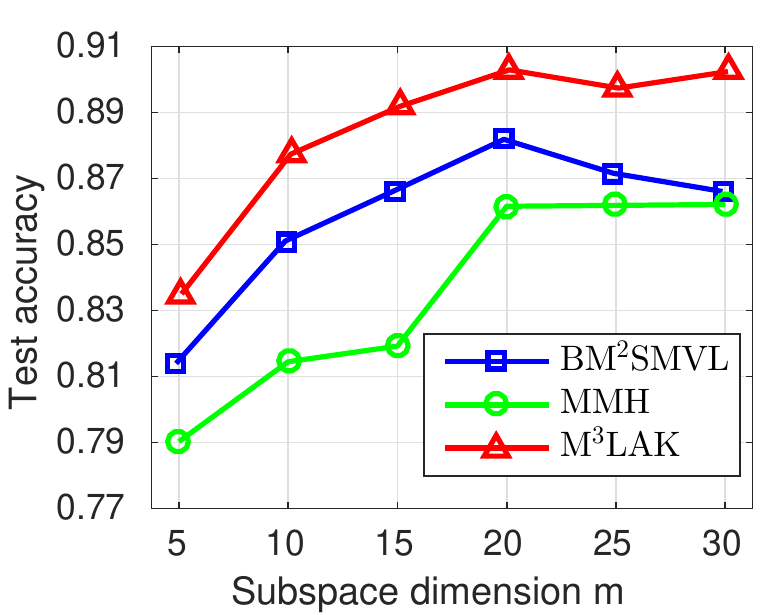}}
\hspace {0.8cm}
\subfigure[Wisconsin] {\includegraphics[height=1.7in,width=2in]{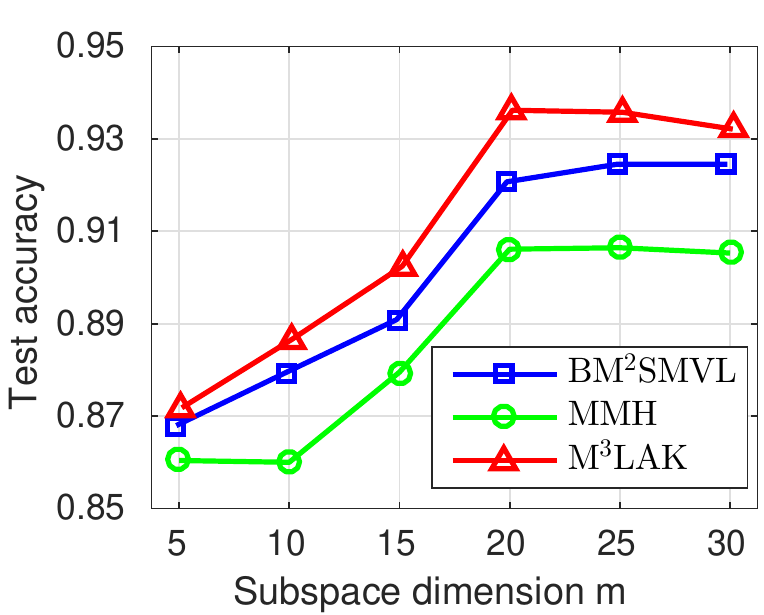}}
\vskip -0.01in
\caption{Effect of subspace dimension $m$. }
\vskip -0.1in
\end{figure}
\vspace{-0.1cm}
\begin{figure}[!htbp]
\centering
\subfigure[Effect of $C$] {\includegraphics[height=1.7in,width=2in]{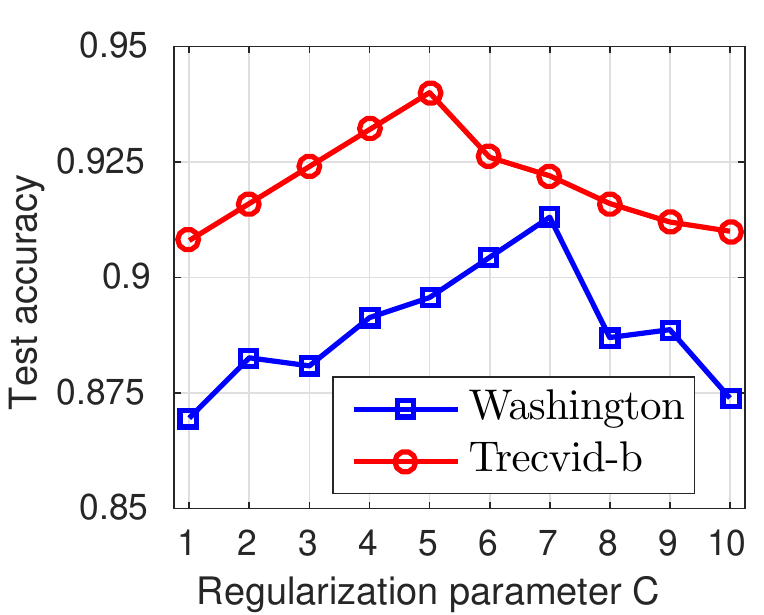}}
\hspace {0.8cm}
\subfigure[Convergency] {\includegraphics[height=1.7in,width=2in]{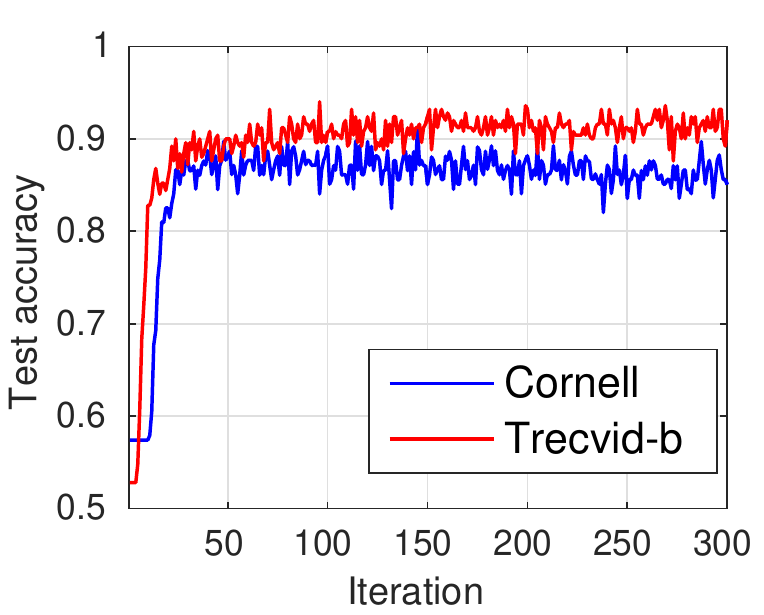}}
\vskip -0.01in
\caption{(a) Effect of regularization parameter $C$ in M$^3$LAK; (b) Convergency of M$^3$LAK. }
\vskip -0.1in
\end{figure}

\vspace{-0.1cm}
\section{Related Work}
\vspace{-0.1cm}
The earliest works of multi-view learning are introduced by Blum and Mitchell~\cite{blum1998combining} and Yarowsky~\cite{yarowsky1995unsupervised}.
Nowadays, a large number of studies are devoted to multi-view learning, e.g., multiple kernel learning~\cite{gonen2011multiple}, disagreement-based multi-view learning~\cite{blum1998combining} and late fusion methods which combine outputs of the models constructed from different view features~\cite{ye2012robust}. Though some linear multi-view learning methods~\cite{Xu2014Large,heonline} are reported to have a good performance on many problems, these methods have made linearity assumption on the data, which may be inappropriate for multi-view data revealing nonlinearities.

As is well known, Kernel method is a principled way for introducing nonlinearity into linear models. So many single kernel nonlinear multi-view learning methods have been proposed~\cite{farquhar2005two,Szedmak2007Synthesis,Fang2012Simultaneously,sun2013multi,quang2013unifying}, which typically require the user to select and tune a predefined kernel for each view. Choosing an appropriate kernel for real-world situations is usually not easy for users without enough domain knowledge. The performance of these models may be greatly affected by the choice of
kernel. An alternative solution to resolve this problem is provided by multiple kernel learning (MKL), which can predefine different kernels for each data view and then integrate the kernels by algorithms such as SDP~\cite{lanckriet2004learning}, SILP~\cite{Sonnenburg2006Large}, simple MKL~\cite{rakotomamonjy2008simplemkl} and BEMKL~\cite{Gonen2012Bayesian}.
However, MKL models inherently have to compute and store many Gram matrices to get good performance while computing a Gram matrix needs $O(N^2D)$ operations and storing too many Gram matrices often leads to out of memory on commonly used computers.

Besides, a nonlinear support vector machines is proposed by using a Gaussian process prior~\cite{Henao2014Bayesian}. But the time complexity for computing the inverse of a covariance matrix is $O(N^3)$.

\section{Conclusion}
\vspace{-0.1cm}
In this paper, we present an adaptive kernel nonlinear max-margin multi-view learning framework. It regularizes the posterior of an efficient multi-view LVM by explicitly mapping the latent representations extracted from multiple data views to a random fourier feature space where max-margin classification constraints are imposed. Having no need to compute the Gram matrix, the computational complexity of our algorithm is linear w.r.t. $N$. Extensive experiments on real-world datasets demonstrate our method has a superior performance, compared with a number of competitors.

\smallskip
\section{Acknowledgments}
\vspace{-0.1cm}
This work was supported by the National Natural Science Foundation of China (No. 61602449, 61473273, 91546122, 61573335, 61602438, 61473274), National High-tech R\&D
Program of China (863 Program) (No.2014AA015105),
Guangdong provincial science and technology plan projects
(No. 2015 B010109005), the Youth Innovation Promotion
Association CAS 2017146.

\bibliographystyle{plain}
\bibliography{M3LAK}

\begin{thebibliography}{10}

\bibitem{blum1998combining}
Avrim Blum and Tom Mitchell.
\newblock Combining labeled and unlabeled data with co-training.
\newblock In {\em COLT}, pages 92--100, 1998.

\bibitem{chen2012large}
Ning Chen, Jun Zhu, Fuchun Sun, and Eric~Poe Xing.
\newblock Large-margin predictive latent subspace learning for multiview data
  analysis.
\newblock {\em PAMI}, 34(12):2365--2378, 2012.

\bibitem{nus-wide-civr09}
Tat-Seng Chua, Jinhui Tang, Richang Hong, Haojie Li, Zhiping Luo, and Yan-Tao.
  Zheng.
\newblock Nus-wide: A real-world web image database from national university of
  singapore.
\newblock In {\em CIVR}, Santorini, Greece., July 8-10, 2009.

\bibitem{Du2016Online}
Changying Du, Changde Du, Guoping Long, Qing He, and Yucheng Li.
\newblock Online bayesian multiple kernel bipartite ranking.
\newblock In {\em Conference on Uncertainty in Artificial Intelligence}, 2016.

\bibitem{Du2016Efficient}
Changying Du, Changde Du, Guoping Long, Jin Xin, and Yucheng Li.
\newblock Efficient bayesian maximum margin multiple kernel learning.
\newblock In {\em ECML-PKDD}, 2016.

\bibitem{Du2015Bayesian}
Changying Du, Shandian Zhe, Fuzhen Zhuang, Qi~Yuan, and Zhongzhi Shi.
\newblock Bayesian maximum margin principal component analysis.
\newblock In {\em Twenty-ninth AAAI Conference on Artificial Intelligence},
  2015.

\bibitem{Fang2012Simultaneously}
Zheng Fang and Zhongfei Zhang.
\newblock Simultaneously combining multi-view multi-label learning with maximum
  margin classification.
\newblock In {\em ICDM}, pages 864--869, 2012.

\bibitem{farquhar2005two}
Jason Farquhar, David Hardoon, Hongying Meng, John~S Shawe-taylor, and Sandor
  Szedmak.
\newblock Two view learning: Svm-2k, theory and practice.
\newblock In {\em Advances in neural information processing systems}, pages
  355--362, 2005.

\bibitem{ge2015distributed}
Hong Ge, Yutian Chen, ENG CAM, Moquan Wan, and Zoubin Ghahramani.
\newblock Distributed inference for dirichlet process mixture models.
\newblock In {\em ICML}, pages 2276--2284, 2015.

\bibitem{ghosh2003bayesian}
Jayanta~K Ghosh and RV~Ramamoorthi.
\newblock {\em Bayesian nonparametrics}, volume~1.
\newblock Springer New York, 2003.

\bibitem{Gonen2012Bayesian}
Mehmet Gonen.
\newblock Bayesian efficient multiple kernel learning.
\newblock In {\em ICML}, pages 1--8, 2012.

\bibitem{gonen2011multiple}
Mehmet G{\"o}nen and Ethem Alpayd{\i}n.
\newblock Multiple kernel learning algorithms.
\newblock {\em JMLR}, 12:2211--2268, 2011.

\bibitem{heonline}
Jia He, Changying Du, Fuzhen Zhuang, Xin Yin, Qing He, and Guoping Long.
\newblock Online bayesian max-margin subspace multi-view learning.
\newblock {\em IJCAI}, pages 1555--1561, 2016.

\bibitem{Henao2014Bayesian}
Ricardo Henao, Xin Yuan, and Lawrence Carin.
\newblock Bayesian nonlinear support vector machines and discriminative factor
  modeling.
\newblock In {\em Neural Information Processing Systems}, pages 1754--1762,
  2014.

\bibitem{Hofmann2007Kernel}
Thomas Hofmann, Bernhard Sch\"{o}lkopf, and Alexander~J. Smola.
\newblock Kernel methods in machine learning.
\newblock {\em Annals of Statistics}, 36(3):1171--1220, 2007.

\bibitem{jaakkola1999maximum}
Tommi Jaakkola, Marina Meila, and Tony Jebara.
\newblock Maximum entropy discrimination.
\newblock In {\em Advances in neural information processing systems}, 1999.

\bibitem{kalli2011slice}
Maria Kalli, Jim~E Griffin, and Stephen~G Walker.
\newblock Slice sampling mixture models.
\newblock {\em Statistics and computing}, 21(1):93--105, 2011.

\bibitem{lanckriet2004learning}
Gert~RG Lanckriet, Nello Cristianini, Peter Bartlett, Laurent~El Ghaoui, and
  Michael~I Jordan.
\newblock Learning the kernel matrix with semidefinite programming.
\newblock {\em JMLR}, 5:27--72, 2004.

\bibitem{neal2011mcmc}
Radford~M Neal.
\newblock Mcmc using hamiltonian dynamics.
\newblock {\em Handbook of Markov Chain Monte Carlo}, 2:113--162, 2011.

\bibitem{Oliva2016Bayesian}
Junier Oliva, Avinava Dubey, Barnabas Poczos, Jeff Schneider, and Eric~P. Xing.
\newblock Bayesian nonparametric kernel learning.
\newblock {\em AISTATS}, 2016.

\bibitem{polson2011data}
Nicholas~G Polson and Steven~L Scott.
\newblock Data augmentation for support vector machines.
\newblock {\em Bayesian Analysis}, 6(1):1--23, 2011.

\bibitem{quang2013unifying}
Minh~H Quang, Loris Bazzani, and Vittorio Murino.
\newblock A unifying framework for vector-valued manifold regularization and
  multi-view learning.
\newblock In {\em ICML}, pages 100--108, 2013.

\bibitem{rahimi2007random}
Ali Rahimi and Benjamin Recht.
\newblock Random features for large-scale kernel machines.
\newblock In {\em Advances in neural information processing systems}, pages
  1177--1184, 2007.

\bibitem{rakotomamonjy2008simplemkl}
Alain Rakotomamonjy, Francis~R Bach, St{\'e}phane Canu, and Yves Grandvalet.
\newblock Simplemkl.
\newblock {\em JMLR}, 9(Nov):2491--2521, 2008.

\bibitem{rudin2011fourier}
Walter Rudin.
\newblock {\em Fourier analysis on groups}.
\newblock John Wiley \& Sons, 2011.

\bibitem{Sonnenburg2006Large}
S\"{o}ren Sonnenburg, Gunnar R\"{a}tsch, Christin Sch\"{a}fer, and Bernhard
  Sch\"{o}lkopf.
\newblock Large scale multiple kernel learning.
\newblock {\em JMLR}, 7(2006):1531--1565, 2006.

\bibitem{sun2013multi}
Shiliang Sun and Guoqing Chao.
\newblock Multi-view maximum entropy discrimination.
\newblock In {\em IJCAI}, pages 1706--1712, 2013.

\bibitem{Szedmak2007Synthesis}
Sandor Szedmak and John Shawe-Taylor.
\newblock Synthesis of maximum margin and multiview learning using unlabeled
  data.
\newblock {\em Neurocomputing}, 70(7-9):1254--1264, 2007.

\bibitem{walker2007sampling}
Stephen~G Walker.
\newblock Sampling the dirichlet mixture model with slices.
\newblock {\em Communications in Statistics-Simulation and
  Computation{\textregistered}}, 36(1):45--54, 2007.

\bibitem{wang2007variational}
Chong Wang.
\newblock Variational bayesian approach to canonical correlation analysis.
\newblock {\em Neural Networks}, 18(3):905--910, 2007.

\bibitem{Xu2014Large}
Chang Xu, Dacheng Tao, Yangxi Li, and Chao Xu.
\newblock Large-margin multi-view gaussian process.
\newblock {\em Multimedia Systems}, 21(2):147--157, 2014.

\bibitem{yarowsky1995unsupervised}
David Yarowsky.
\newblock Unsupervised word sense disambiguation rivaling supervised methods.
\newblock In {\em ACL}, pages 189--196, 1995.

\bibitem{ye2012robust}
Guangnan Ye, Dong Liu, I-Hong Jhuo, Shih-Fu Chang, et~al.
\newblock Robust late fusion with rank minimization.
\newblock In {\em CVPR}, pages 3021--3028, 2012.

\bibitem{zhu2012medlda}
Jun Zhu, Amr Ahmed, and Eric~P Xing.
\newblock Medlda: maximum margin supervised topic models.
\newblock {\em JMLR}, 13(1):2237--2278, 2012.

\bibitem{Zhuang2012Multi}
Fuzhen Zhuang, George Karypis, Xia Ning, Qing He, and Zhongzhi Shi.
\newblock Multi-view learning via probabilistic latent semantic analysis.
\newblock {\em Information Sciences}, 199(15):20--30, 2012.

\end{thebibliography}

\end{document}